\documentclass[conference]{IEEEtran}
\IEEEoverridecommandlockouts
\usepackage{cite}
\usepackage{amsmath,amssymb,amsfonts}
\usepackage{algorithmic}
\usepackage{graphicx}
\usepackage{textcomp}
\usepackage{bm}
\usepackage{booktabs}
\usepackage{xcolor}
\usepackage{url}
\def\BibTeX{{\rm B\kern-.05em{\sc i\kern-.025em b}\kern-.08em
    T\kern-.1667em\lower.7ex\hbox{E}\kern-.125emX}}
\begin{document}

\title{DroneIQA-VLE: Multi-Task Drone Image Quality Assessment via Vision-Language Ensemble}
\author{Wei Sun\textsuperscript{1}, Weixia Zhang\textsuperscript{2}, Hongjian Zhan\textsuperscript{1}, Mingkai Lu\textsuperscript{2}, Yixuan Gao\textsuperscript{2}, Guangtao Zhai\textsuperscript{2}\\
\textsuperscript{1}East China Normal University, \textsuperscript{2}Shanghai Jiao Tong University}
\maketitle

\begin{abstract}
We present DroneIQA-VLE, our solution to the ICME 2026 Drone-IQA Grand Challenge on Target-aware Image Quality Assessment for Low-altitude UAV Images. The framework jointly predicts global, target, and background quality scores by ensembling two complementary pipelines: (1) SigLIP2 vision encoders with multi-task regression heads, and (2) a LoRA-adapted Qwen3.5-9B multimodal large language model for quality score regression. The final global quality prediction is obtained by arithmetically averaging the outputs of both pipelines. Our method achieves 2nd place in the challenge, demonstrating its effectiveness. The code is available at \url{https://github.com/sunwei925/DroneIQA-VLE}.

\end{abstract}

\section{Introduction}
Unmanned Aerial Vehicle (UAV) imagery has become increasingly prevalent in applications such as surveillance, traffic monitoring, and emergency response. However, UAV images exhibit distinct quality characteristics compared to conventional natural images due to diverse viewpoints, small target regions, complex backgrounds, and spatially nonuniform degradations. These factors make standard image quality assessment (IQA)~\cite{zhai2020perceptual} methods less suitable for UAV scenarios. Traditional full-reference metrics such as PSNR and SSIM~\cite{wang2004image} are impractical since pristine references are unavailable in real-world UAV deployments, while most no-reference IQA methods~\cite{zhang2023blind,sun2023blind,lu2022deep,wang2024large,sun2022deep,cao2026vqathinker,sun2025efficient} focus solely on global perceptual quality without considering target-region usability and background interference.

To promote research on UAV-oriented quality modeling, the Drone-IQA GC 2026 Grand Challenge~\cite{jiang2026overview} introduces a target-aware benchmark comprising approximately 6,000 UAV images collected from VisDrone~\cite{zhu2018vision} and UAVDT~\cite{du2018unmanned} datasets, annotated by 18 human raters along three perceptual dimensions: global quality, target quality, and background quality. The challenge requires participants to predict the global quality score of UAV images, while target and background quality annotations can serve as auxiliary supervision. Submissions are evaluated using the average of Pearson Linear Correlation Coefficient (PLCC) and Spearman Rank Correlation Coefficient (SRCC).

In this report, we present DroneIQA-VLE, our solution to this challenge. Our approach is motivated by two key observations. First, multi-task learning with auxiliary quality dimensions (target and background) provides beneficial inductive bias for the primary global quality prediction task. Second, vision-encoder-based models and multimodal large language models capture complementary quality-relevant features: the former excels at spatial quality feature extraction, while the latter provides high-level semantic understanding and perceptual reasoning. By ensembling predictions from SigLIP2 vision encoders and a LoRA-adapted Qwen3.5-9B model, our framework achieves robust and accurate quality predictions that generalize well across diverse UAV imaging conditions.



\section{Model Architecture}
\label{sec:architecture}

We investigate two complementary modeling paradigms for multi-task drone image quality assessment: pure vision-encoder regression and multimodal large language model regression.

\subsection{SigLIP2 Multi-task Models}
\label{sec:siglip2}

We employ two SigLIP2-based visual backbones spanning different architectural scales within the Vision Transformer family. Each backbone is equipped with three independent regression heads to predict the three quality dimensions simultaneously.

\textbf{SigLIP2 ViT-L/16 (384\,px).}
The first architecture builds upon the visual encoder of SigLIP2 ViT-L/16~\cite{tschannen2025siglip}, a Vision Transformer Large model with a patch size of 16 that operates at a $384 \times 384$ input resolution. It produces a 1024-dimensional feature vector per input image. Three lightweight quality regression heads, each consisting of two cascaded fully connected layers ($1024 \!\to\! 128$ and $128 \!\to\! 1$) without intermediate nonlinearities, are appended to project the feature representation onto scalar quality scores for the three tasks.

\textbf{SigLIP2 ViT-SO400M/14 (378\,px).}
The second architecture employs the SigLIP2 ViT-SO400M/14~\cite{tschannen2025siglip}, a larger-capacity vision transformer with a patch size of 14, operating at a native resolution of $378 \times 378$ ($378 = 27 \times 14$). Dynamic image size support is enabled to allow flexible resolution inputs. This model produces a 1152-dimensional feature vector per image. Three quality regression heads of analogous design ($1152 \!\to\! 128$ and $128 \!\to\! 1$) map the extracted representations to scalar quality predictions.

In both architectures, the three regression heads predict \texttt{global\_quality\_mean}, \texttt{target\_quality\_mean}, and \texttt{background\_quality\_mean} independently, sharing no parameters with one another beyond the common visual backbone.

\subsection{Qwen3.5-9B Multimodal LLM}
\label{sec:qwen}

The second pipeline is based on the Qwen3.5-9B multimodal large language model~\cite{team2026qwen3}, adapted for quality regression. Given an input drone image and a task-specific prompt, the model learns a multimodal representation that captures quality-relevant features at both semantic and perceptual levels. Instead of generating textual outputs, the framework directly uses the hidden representations of the MLLM as high-level features and employs a regression head to simultaneously predict three continuous quality scores corresponding to global quality, target quality, and background quality.

For efficient adaptation, LoRA-based fine-tuning~\cite{hu2022lora} is adopted with rank 64 and scaling factor $\alpha = 128$, applied to all linear modules across the visual encoder, the visual aligner, and the large language model, while keeping most pretrained parameters frozen. The model is trained under bfloat16 mixed precision with a maximum prompt length of 4096 tokens and a maximum of 2048 image tokens.

\subsection{Ensemble Strategy}
\label{sec:ensemble}

The final prediction is obtained by averaging the \texttt{global\_quality} predictions from both the SigLIP2 ensemble and the Qwen3.5-9B model:
\begin{equation}
    \hat{Q} = \frac{Q_{\text{SigLIP2}} + Q_{\text{Qwen}}}{2},
    \label{eq:ensemble}
\end{equation}
where $Q_{\text{SigLIP2}}$ denotes the averaged global quality prediction across all SigLIP2 checkpoints, and $Q_{\text{Qwen}}$ denotes the Qwen3.5-9B prediction. This cross-architecture ensemble leverages complementary strengths: the SigLIP2 models provide efficient, multi-scale vision-encoder-based quality features with strong spatial quality awareness, while Qwen3.5-9B contributes rich multimodal reasoning and high-level semantic understanding capabilities.

\section{Training Procedure}
\label{sec:training}

\subsection{Dataset}
We train and evaluate our model on the Drone-IQA GC 2026 benchmark~\cite{droneiqa2026}. The challenge provides 3,600 images with released annotations as the training set, 1,200 images as the validation set, and a held-out test set for final evaluation. Each image is annotated with three quality dimensions: global quality, target quality, and background quality. The final ranking is determined by the average of PLCC and SRCC on the global quality prediction.

\subsection{Image Preprocessing}
\label{sec:preprocess}

\textbf{SigLIP2.}
Each image is first resized such that its shorter side equals 432 pixels, preserving the original aspect ratio. During training, a random crop of $384 \times 384$ (for ViT-L/16) or $378 \times 378$ (for ViT-SO400M/14) is extracted. During inference, a deterministic center crop of the same size is applied to ensure reproducibility. All pixel values are normalized with $\text{mean} = [0.5, 0.5, 0.5]$ and $\text{std} = [0.5, 0.5, 0.5]$.

\textbf{Qwen3.5-9B.}
Images are processed through the Qwen3.5-9B built-in visual preprocessing pipeline. The maximum number of image tokens is capped at 2048 and the maximum prompt sequence length is set to 4096 tokens.

\subsection{Training Details --- SigLIP2}
\label{sec:train_siglip}

Both SigLIP2 architectures are trained with an identical protocol. The dataset is randomly split into 80\% training and 20\% validation, with 3 independent splits (controlled by different random seeds) to reduce variance from data-dependent biases.

All models are optimized using Adam~\cite{kingma2014adam} with an initial learning rate of $1 \times 10^{-5}$ and a weight decay of $1 \times 10^{-7}$. The learning rate is decayed by a factor of 0.95 every 2 epochs following a step-wise schedule. The training loss is the sum of PLCC losses across all three tasks~\cite{sun2025enhancing}:
\begin{equation}
    \mathcal{L} = \sum_{t \in \mathcal{T}} \mathcal{L}_{\text{PLCC}}^{(t)},
    \label{eq:total_loss_siglip}
\end{equation}
where $\mathcal{T}$ is a set including the dimensions of \texttt{global\_quality}, \texttt{target\_quality}, and \texttt{background\_quality}.

Training is conducted for 10 epochs with a batch size of 32. The best checkpoint for each split is selected based on the highest $(\text{SRCC} + \text{PLCC})/2$ computed on the \texttt{global\_quality} validation metric.

\subsection{Training Details --- Qwen3.5-9B}
\label{sec:train_qwen}

The Qwen3.5-9B model is fine-tuned using the ms-swift framework with LoRA adaptation. The task is configured as sequence classification with 3 regression labels. The model is trained with a combination of a PLCC-induced loss and a fidelity loss~\cite{sun2024assessing,zhang2023blind}:
\begin{equation}
    \mathcal{L} = \lambda_1 \mathcal{L}_{\text{Fid}} + \lambda_2 \mathcal{L}_{\text{PLCC}}.
    \label{eq:total_loss_qwen}
\end{equation}

Given a mini-batch of predictions $\hat{y}_i$ and ground-truth scores $y_i$, where $i = 1, \ldots, B$ indexes samples, the PLCC-induced loss is defined as:
\begin{equation}
    \mathcal{L}_{\text{PLCC}} = 1 - \frac{\displaystyle\sum_{i=1}^{B}(\hat{y}_i - \bar{\hat{y}})(y_i - \bar{y})}{\displaystyle\sqrt{\sum_{i=1}^{B}(\hat{y}_i - \bar{\hat{y}})^2}\;\sqrt{\sum_{i=1}^{B}(y_i - \bar{y})^2}},
    \label{eq:plcc_loss_qwen}
\end{equation}
where $\bar{\hat{y}} = \frac{1}{B}\sum_{i=1}^{B}\hat{y}_i$ and $\bar{y} = \frac{1}{B}\sum_{i=1}^{B}y_i$. This term encourages high linear correlation between predictions and subjective scores.

The fidelity loss is formulated on pairwise score differences:
\begin{equation}
    \mathcal{L}_{\text{Fid}} = \frac{1}{|\mathcal{P}|}\!\sum_{(i,j)\in\mathcal{P}}\!\left[1 - \left(\sqrt{p_{ij}\,g_{ij}} + \sqrt{(1-p_{ij})(1-g_{ij})}\right)\right],
    \label{eq:fid_loss}
\end{equation}
where
\begin{equation}
    p_{ij} = \Phi(\hat{y}_i - \hat{y}_j), \quad
    g_{ij} = \frac{\text{sign}(y_i - y_j) + 1}{2},
    \label{eq:fid_terms}
\end{equation}
and $\mathcal{P} = \{(i,j) \mid i < j\}$. Here, $\Phi(\cdot)$ denotes the standard Gaussian cumulative distribution function. This term enforces consistency between the predicted pairwise ordering and the ground-truth ordering.

The initial learning rate is set to $1 \times 10^{-4}$ and updated with a cosine decay schedule. Training is conducted for 3 epochs with a per-GPU batch size of 16 under bfloat16 precision.

\subsection{Inference and Ensemble}
\label{sec:inference}

At inference time, we employ a cross-pipeline ensemble to improve prediction robustness. The SigLIP2 pipeline loads 6 checkpoints (2 architectures $\times$ 3 splits) and computes per-image quality predictions by averaging across all models after applying a four-parameter logistic mapping fitted on the corresponding validation splits. The Qwen3.5-9B pipeline independently processes each image through the LoRA-adapted model and applies logistic mapping using pre-fitted parameters stored in a separate parameter file.

The final global quality prediction is obtained by arithmetic averaging of the \texttt{global\_quality} scores from both pipelines. This strategy leverages two complementary sources of diversity: the SigLIP2 models capture multi-scale spatial quality features through their vision-encoder architectures, while the Qwen3.5-9B model provides high-level multimodal reasoning informed by large-scale language--vision pretraining, leading to more accurate and generalizable quality predictions.

\section{Experiment Results}

\begin{table}[t]
\centering
\caption{Experiments Results of the ICME 2026 Drone-IQA Grand Challenge.}
\label{tab:experiment}
\begin{tabular}{c l c c c}
\toprule
Rank & Team & PLCC & SRCC & Score \\
\midrule
1 & cmsr & 0.9512 & 0.9450 & 0.9481 \\
2 & VQA (DroneIQA-VLE) & 0.9484 & 0.9420 & 0.9452 \\
3 & Echo & 0.9394 & 0.9332 & 0.9363 \\
4 & TASEAI & 0.9293 & 0.9244 & 0.9268 \\
5 & Watrix & 0.9262 & 0.9226 & 0.9244 \\
\bottomrule
\end{tabular}
\end{table}

Table~\ref{tab:experiment} presents the final results on the held-out test set as evaluated by the challenge organizers. Our method (Team VQA) ranks 2nd among all participating teams, demonstrating the effectiveness of our vision-language ensemble strategy. Compared with the 3rd-place method, our approach achieves a notable improvement of approximately 0.9 percentage points in the overall score, validating the complementary benefits of combining SigLIP2-based spatial quality features with Qwen3.5-9B's multimodal reasoning capabilities. Moreover, the narrow gap relative to the 1st-place team (less than 0.3 percentage points) further confirms the competitiveness of our approach. Both PLCC and SRCC remain consistently high, indicating that our predictions are well-calibrated in terms of both linear correlation and rank-order agreement with human subjective judgments.

\section{Conclusion}

In this report, we present DroneIQA-VLE, a vision–language ensemble framework for target-aware UAV image quality assessment. The framework integrates two complementary pipelines: SigLIP2 vision encoders with multi-task regression heads and a LoRA-adapted Qwen3.5-9B multimodal large language model. By ensembling the predictions from both pipelines, our method effectively captures diverse quality-aware representations and achieves second place in the ICME 2026 Drone-IQA Grand Challenge, demonstrating its effectiveness.


\bibliographystyle{IEEEbib}
\bibliography{icme2025references}

\vspace{12pt}

\end{document}